\documentclass[10pt,twocolumn,letterpaper]{article}

\usepackage{cvpr}              

\definecolor{cvprblue}{rgb}{0.21,0.49,0.74}
\usepackage{multirow}
\usepackage[pagebackref,breaklinks,colorlinks,allcolors=cvprblue]{hyperref}
\usepackage{array}

\def\paperID{32524} 
\def\confName{CVPR}
\def\confYear{2026}

\title{FoV-Net: Rotation-Invariant CAD B-rep Learning via Field-of-View Ray Casting}

\author{Matteo Ballegeer\\
Ghent University\\
CVAMO, FlandersMake@UGent\\
{\tt\small matteo.ballegeer@ugent.be}
\and
Dries F. Benoit\\
Ghent University\\
CVAMO, FlandersMake@UGent\\
{\tt\small dries.benoit@ugent.be}
}

\begin{document}
\maketitle
\begin{abstract}
Learning directly from boundary representations (B-reps) has significantly advanced 3D CAD analysis. 
However, state-of-the-art B-rep learning methods rely on absolute coordinates and normals to encode global context, making them highly sensitive to rotations. 
Our experiments reveal that models achieving over 95\% accuracy on aligned benchmarks can collapse to as low as 10\% under arbitrary $\mathbf{SO}(3)$ rotations. 
To address this, we introduce FoV-Net, the first B-rep learning framework that captures both local surface geometry and global structural context in a rotation-invariant manner.
Each face is represented by a Local Reference Frame (LRF) UV-grid that encodes its local surface geometry, and by Field-of-View (FoV) grids that capture the surrounding 3D context by casting rays and recording intersections with neighboring faces.
Lightweight CNNs extract per-face features, which are propagated over the B-rep graph using a graph attention network. 
FoV-Net achieves state-of-the-art performance on B-rep classification and segmentation benchmarks, demonstrating robustness to arbitrary rotations while also requiring less training data to achieve strong results.
Source code is available at \url{https://github.com/UGent-CVAMO/fovnet}
\end{abstract}    
\section{Introduction}
\label{sec:intro}
Modern manufacturing increasingly relies on Computer-Aided Design (CAD), driving demand for 3D deep learning methods for part classification, segmentation, and generative design~\cite{jayaraman2021uvnet,lambourne2021brepnet,xu2024brepgen}.
CAD boundary representations (B-reps) naturally form graphs, with faces as nodes and shared edges as connections, motivating the use of graph neural networks (GNNs) directly on B-reps~\cite{jayaraman2021uvnet}.
By encoding topology through the graph and geometry via node and edge features, B-rep learning methods have shown improved performance over point cloud and mesh-based approaches~\cite{jayaraman2021uvnet}.
A key challenge, however, lies in designing descriptors that capture both local surface geometry and structural context.
Many existing descriptors focus on local properties such as curvature, surface type, or area, but fail to capture relationships between faces—for example, whether two adjacent flat faces form an angle or lie on the same plane.
UV-Net~\cite{jayaraman2021uvnet} addresses this by sampling points and normals in each face's parametric UV domain and storing absolute coordinates and normal vectors in the global frame.
This enables descriptors to encode both local surface geometry and global position and orientation, forming the foundation for many recent B-rep learning methods~\cite{lee2023brepgat,wu2024aagnet,xu2024brepgen}.
\begin{figure}[t]
  \centering
  \includegraphics[width=0.8\linewidth]{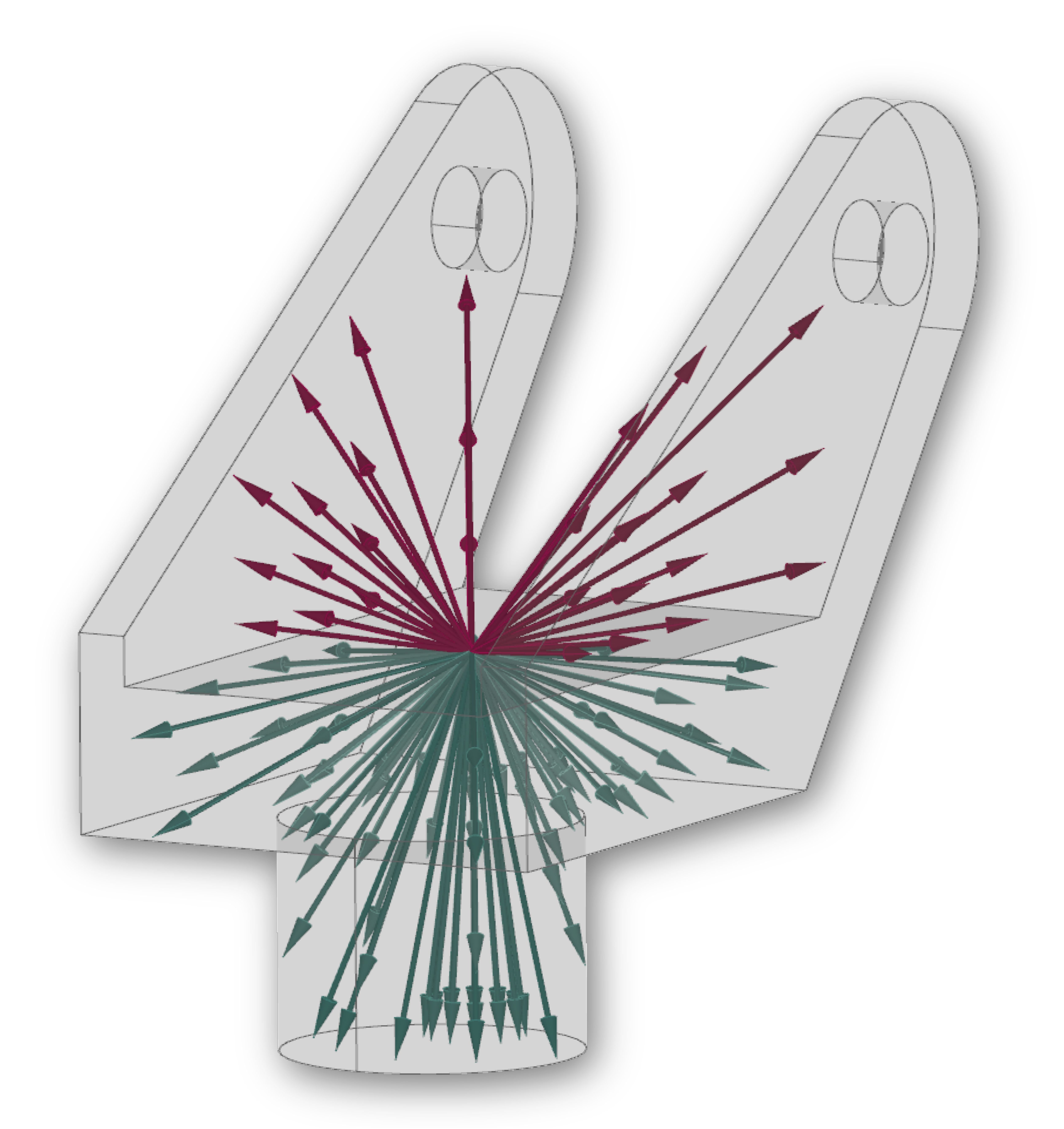}
  \caption{Field-of-View raycasting. 
  Rays are cast from each face center in both inward and outward hemispheres, recording first-hit statistics to capture the surrounding 3D structure from a face-centric viewpoint. 
  Only ray hits are visualized for clarity.}
  \label{img:raycasting}
\end{figure}

Despite strong performance on well-aligned B-rep benchmarks, reliance on global coordinates and normals makes UV-based descriptors inherently sensitive to rotation. 
While a recognized limitation in the original UV-Net work, it has remained unaddressed in subsequent B-rep learning methods.
Our experiments show that accuracy can drop from over 95\% on the original data to as low as 10\% under arbitrary $\mathbf{SO}(3)$ transformations, creating a major limitation for manufacturing pipelines, where CAD models come from diverse sources and analysis must be robust to arbitrary orientations~\cite{wang2023manufacturing}.
Even for aligned models, we show that rotation invariance is crucial in segmentation tasks, as varying face positions can lead to spurious pose correlations that degrade generalization in low-data settings.
Rotation augmentation offers partial mitigation, but cannot cover all rotations in $\mathbf{SO}(3)$ and incurs computational cost, motivating the development of inherently rotation-invariant B-rep descriptors.

We propose FoV-Net, the first B-rep learning framework that captures both local geometry and structural context in a fully rotation-invariant manner.
Our approach draws inspiration from point cloud learning, which emphasizes combining invariant local features with global context~\cite{chen2019clusternet,kim2020rotation,chen2024local}.
Each face is therefore represented by two components: a Local Reference Frame (LRF) UV-grid encoding intrinsic surface geometry, and two Field-of-View (FoV) grids that summarize the surrounding structure.
The FoV grids are generated by casting rays from each face center in the hemisphere along the face normal and its inverse, recording first-hit statistics (\cref{img:raycasting}).
Lightweight CNNs extract per-face features from these grids, which are then propagated across the B-rep graph using a graph attention network (GAT).
FoV-Net achieves strong performance on multiple benchmarks while maintaining full accuracy under arbitrary rotations.
It is also more data-efficient, a valuable trait in industrial CAD settings where intellectual property concerns often plague data availability~\cite{stjepandic2015intellectual}.
In summary:

\begin{itemize}
    \item We provide the first systematic study quantifying rotation sensitivity in B-rep learning.
    \item We propose a rotation-invariant B-rep descriptor that captures both local geometry and surrounding structure.
    \item We introduce FoV-Net, achieving strong, rotation-invariant classification and segmentation performance where rotation-augmented models lose overall accuracy.
    \item We demonstrate superior data efficiency and highlight that rotation sensitivity can degrade face segmentation even on well-aligned data in low-data regimes.
  \end{itemize}
\section{Related Work}
\label{sec:related_work}

\noindent \textbf{Boundary representation learning.}
Boundary representations (B-reps) are the industry standard for encoding 3D solids, representing geometry as parametric surfaces (faces), trimming curves (edges), and explicit topological links~\cite{weiler1986topological,lee2001partial}.
Early deep learning approaches converted CAD models into voxels~\cite{wu20153d}, meshes~\cite{hanocka2019meshcnn}, point clouds~\cite{qi2017pointnet}, or multi-view images~\cite{su2015multi}, often compromising geometric fidelity or topology~\cite{jayaraman2021uvnet}.
Graph-based representations date back to the 1980s as Attributed Adjacency Graphs (AAGs) for feature recognition~\cite{ansaldi1985geometric,joshi1988graph}, where surfaces form graph nodes and shared curves form edges, each with their geometric attributes.
Recent advances in Graph Neural Networks (GNNs) and CAD kernels have revitalized this approach, enabling the encoding of both geometry and topology and achieving strong performance across diverse CAD tasks~\cite{cao2020graph,jayaraman2021uvnet}.
The release of large-scale B-rep datasets has further fueled progress in classification and feature recognition~\cite{cao2020graph,jayaraman2021uvnet,lambourne2021brepnet,lee2023brepgat,zhang2024brepmfr,zou2025bringing}, retrieval~\cite{koch2019abc,jayaraman2021uvnet}, self-supervised pre-training~\cite{jones2023self,lou2023brep}, and generative design~\cite{willis2021fusion,wu2021deepcad,guo2022complexgen,xu2024brepgen,guo2025brepgiff,liu2025hola,li2025dtgbrepgen}.  
A recent survey provides an overview of this growing field~\cite{heidari2025geometric}.

\vspace{0.5em}
\noindent \textbf{Surface and edge descriptors.}
The effectiveness of graph-based B-rep learning methods depends critically on the geometry encoded in per-face and per-edge descriptors.
Early work focused on planar faces, using their implicit surface equations as node features~\cite{cao2020graph}.  
A major advance came with UV-Net~\cite{jayaraman2021uvnet}, which samples each face's UV parameter space to create grid-like arrays of 3D coordinates, normals, and trimming masks.
Processed by CNNs, these grids encode both local surface geometry and global position and orientation.
This UV-based paradigm quickly became standard, with later methods augmenting it with additional attributes such as surface type (plane, cylinder), face area, or edge properties like length and convexity~\cite{colligan2022hierarchical,lee2023brepgat,zhang2024brepmfr}.
Recent methods propose spline or NURBS parameterizations for more efficient surface encodings than UV grids~\cite{fan2024neuronurbs}. 
Crucially, no existing descriptors explicitly capture global structural relationships between B-rep faces beyond UV-based position and orientation.

\vspace{0.5em}
\noindent \textbf{Rotation invariance in 3D Learning.} 
UV-based descriptors are inherently sensitive to rotation because they rely on global coordinates, mirroring challenges long recognized in point cloud learning~\cite{li2021closer,fei2024rotation}.
Proposed solutions include multi-view projections~\cite{su2015multi}, rotation augmentation~\cite{choy2019fully,bai2020d3feat}, PCA-based alignment~\cite{li2021closer}, local reference frames (LRFs)~\cite{chen2019clusternet,chen2024local}, point-pair features~\cite{deng2018ppf}, and invariant descriptors based on relative distances or angles~\cite{zhang2019rotation,li2021rotation}.
Rotation-invariant mesh methods have also been explored~\cite{hanocka2019meshcnn,shakibajahromi2024rimeshgnn}.
Surveys emphasize that only inherently invariant features, such as LRFs, provide guaranteed robustness, since augmentation cannot exhaustively cover all rotations in $\mathbf{SO}(3)$ and is computationally costly~\cite{fei2024rotation}.
Yet, such features often sacrifice global structural context~\cite{zhang2019rotation,yu2023rotation}, motivating the development of descriptors that preserve both local and global information~\cite{kim2020rotation,zhang2020global,zhang2022riconv,yu2023rotation,chen2024local}
While rotation-invariant point cloud descriptors have found applications in CAD manufacturing~\cite{liu2025manufacturing}, B-rep learning still relies on rotation-sensitive UV descriptors to capture global context.
Our method enforces input invariance, whereas promising recent works advocate $\mathbf{SE}(3)$-equivariance (e.g., Vector Neurons~\cite{deng2021vector}, $\mathbf{SE}(3)$-Transformers~\cite{fuchs2020se}, spherical CNNs~\cite{cohen2018spherical}).
These embed rotation equivariance directly into the network for more expressive representations, though requiring specialized features and equivariant layers.
 
\vspace{0.5em}
\noindent \textbf{Ray casting for structural context.}
Ray casting is a classical technique in computer graphics for probing 3D environments~\cite{roth1982ray}.
By emitting rays from a viewpoint and recording first-hit intersections, one can capture information about surrounding geometry.
Beyond traditional applications in visibility analysis and shading, ray casting has been used in 3D geometry processing for tasks such as normal orientation correction and inside/outside determination in meshes~\cite{zhou2008visibility,takayama2014simple} and point clouds~\cite{ochmann2019automatic1,ochmann2019automatic2,fu2025consistent}.
Visibility cones have enabled surface reconstruction from incomplete point clouds~\cite{shalom2010cone}, and learning-based methods have leveraged ray cues for multi-view consistency~\cite{paschalidou2018raynet} or to enhance LiDAR object detection via visibility maps~\cite{hu2020you}.
In industrial CAD, ray casting has been applied to estimate point-wise tool reachability and surface accessibility for machining~\cite{li2012computing}.
These studies demonstrate that global 3D structure can be effectively captured through visibility samples from surface-anchored viewpoints~\cite{li2025mash}.
To our knowledge, we are the first to adapt this idea to B-rep learning, using ray casting to construct rich, rotation-invariant descriptors that encode each face's structural surroundings.
By defining ray origins and directions relative to each face, the resulting descriptors are inherently rotation-invariant, as the viewpoints and rays co-rotate with the face and encode only relative spatial relationships.
\section{Method}
\label{sec:method}
\subsection{Problem formulation}

We represent each CAD model as a B-rep graph $\mathcal{G} = (\mathcal{V}, \mathcal{E})$, where faces are nodes $n_i \in \mathcal{V}$, and edges $(n_i, n_j) \in \mathcal{E}$ connect faces sharing a boundary curve.
Our goal is to learn a per-face feature vector that captures both intrinsic local geometry and structural context in a rotation-invariant manner.
To achieve this, we introduce a novel two-part face descriptor: Local Reference Frame UV (LRF UV) grids capture local surface geometry, and Field-of-View (FoV) grids capture the surrounding structure using ray casting.

\subsection{Input representation}

\begin{figure}
  \centering
  \includegraphics[width=\linewidth]{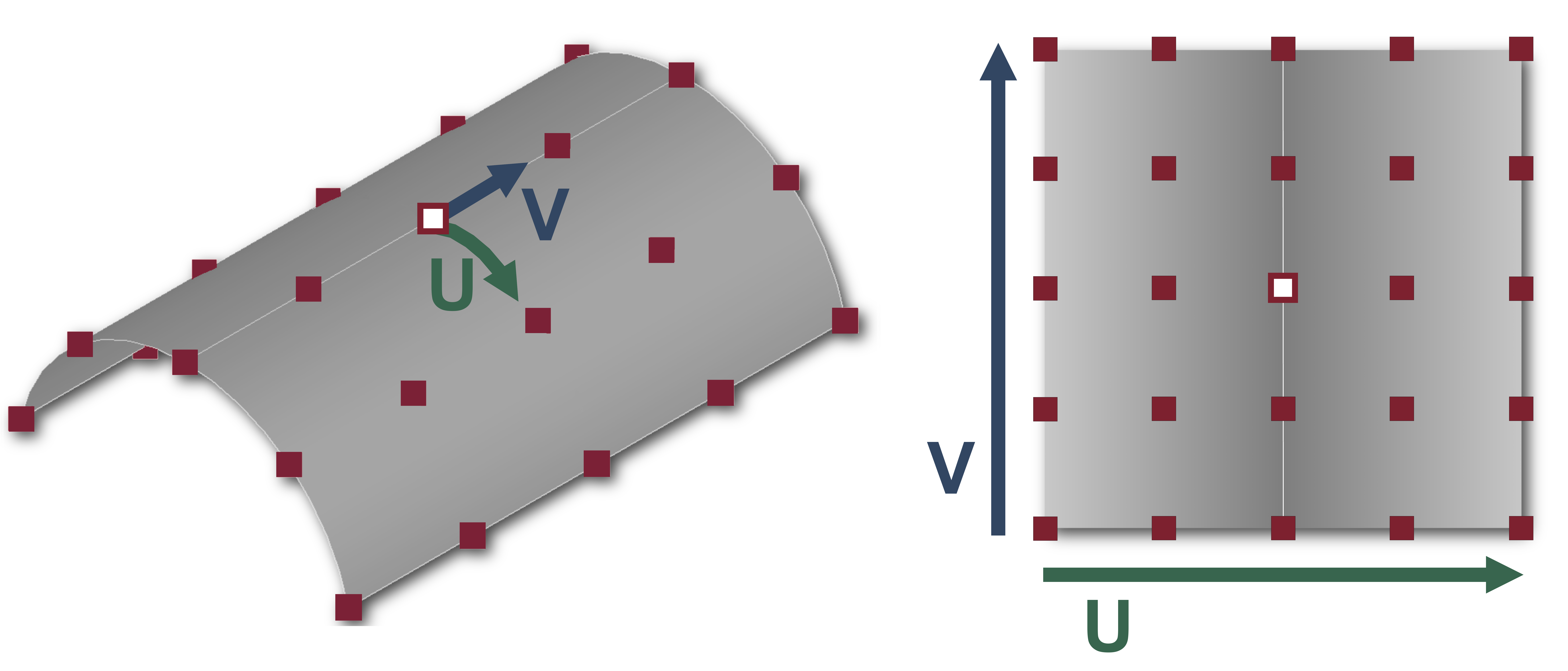}
  \caption{UV-mapping.
  A 3D surface (left) is parameterized by a 2D UV domain (right), where each $(u, v)$ coordinate maps uniquely to a point on the surface. 
  Rasterizing the UV domain produces a fixed-resolution grid of surface samples.}\label{img:uv_grid}
\end{figure}

\begin{figure}[t]
  \centering
  \begin{subfigure}{0.48\linewidth}
    \centering
    \includegraphics[width=\linewidth]{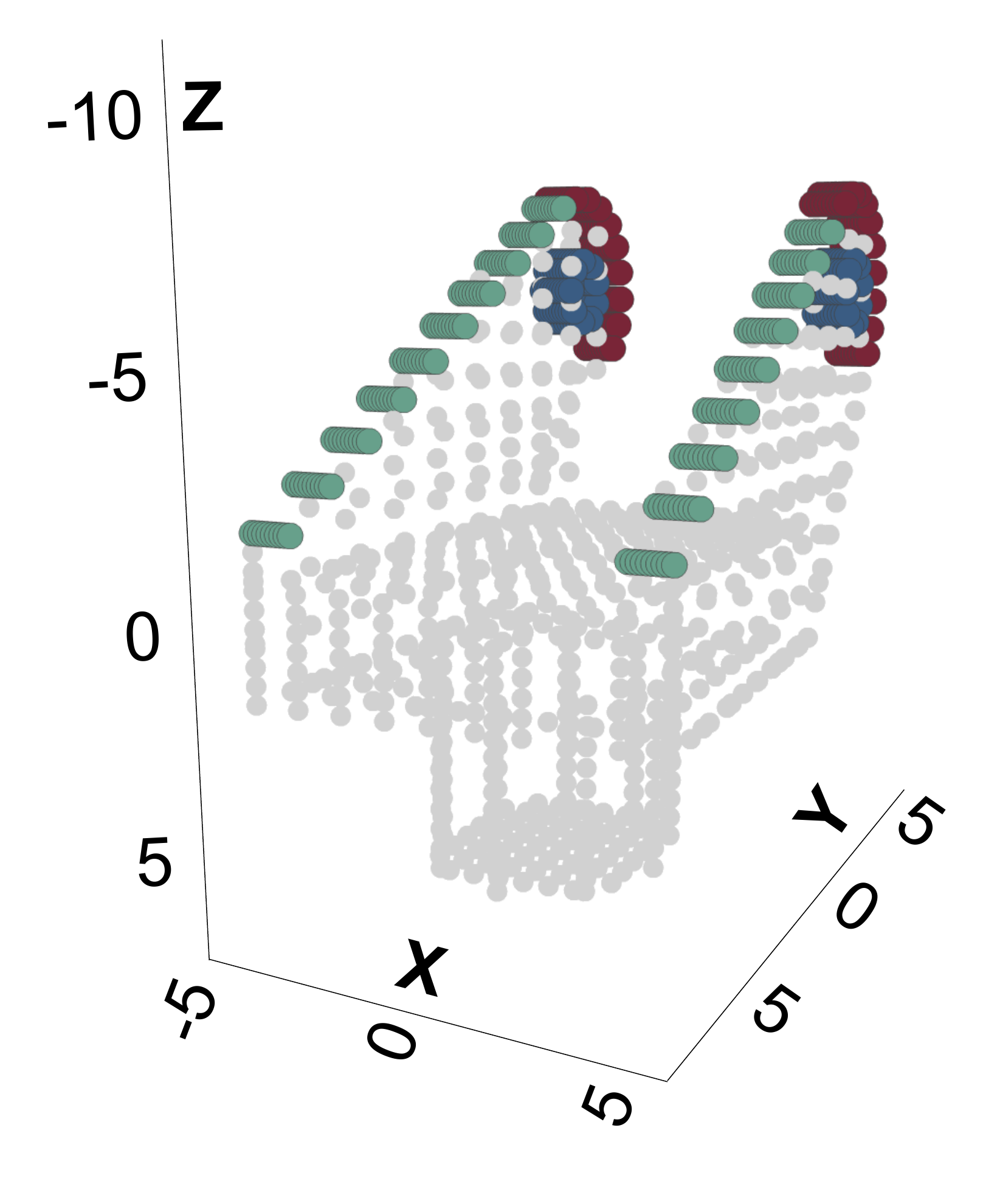}
    \caption{UV-grids in global frame}
    \label{img:global_frame}
  \end{subfigure}
  \hfill
  \begin{subfigure}{0.48\linewidth}
    \centering
    \includegraphics[width=\linewidth]{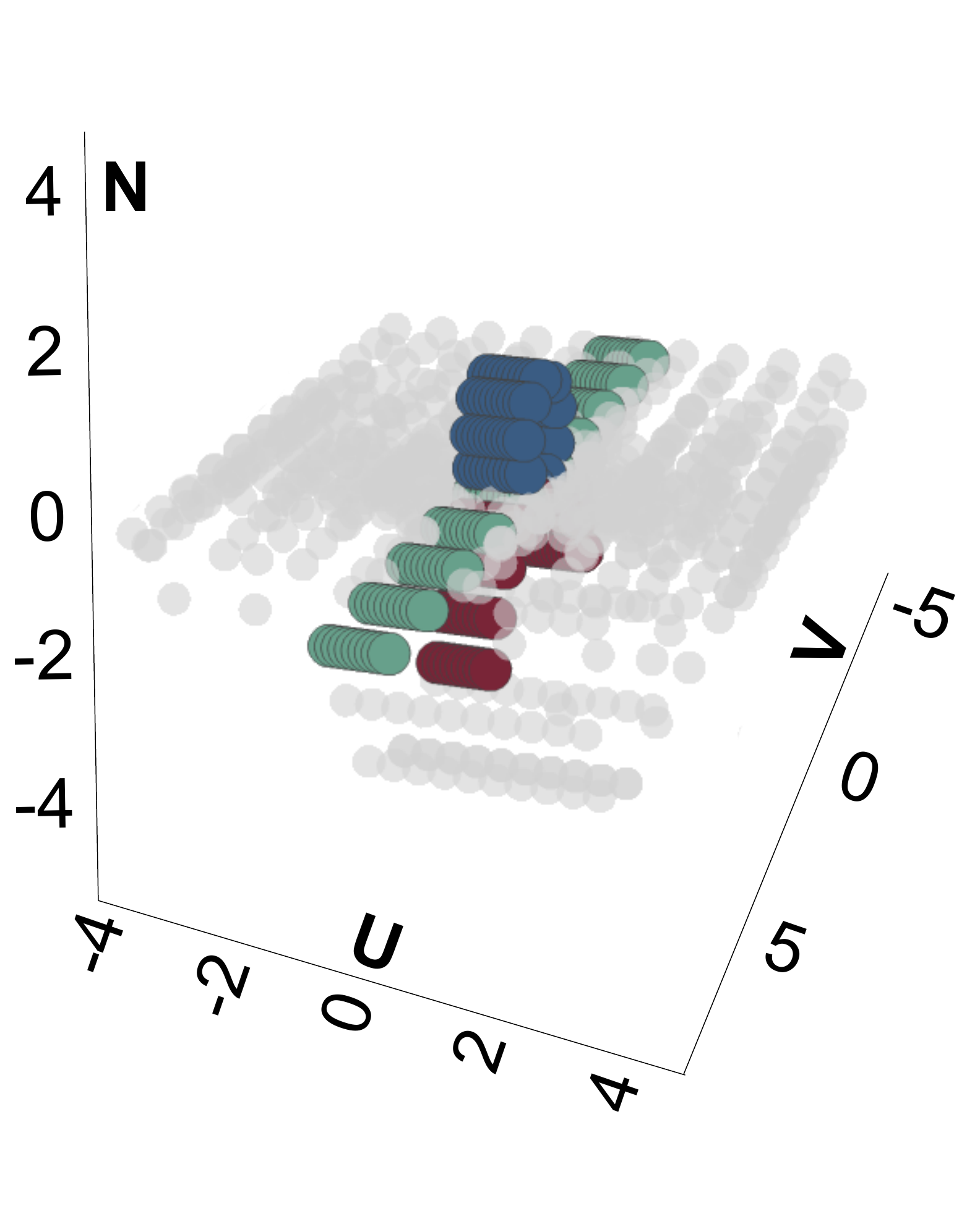}
    \caption{LRF UV-grids in local frame}
    \label{img:local_frame}
  \end{subfigure}
  \caption{LRF UV construction. 
  (a) Standard UV-grids are defined in the global frame (XYZ). 
  (b) LRF UV grids are defined in the $\mathbf{R}_f$ frame (UVN) and relative to $\mathbf{o}$ making identical faces (colors) yield identical descriptors regardless of pose.}\label{img:lrf_uv_construction}
\end{figure}

\noindent \textbf{UV-grids.}
Each B-rep surface can be parameterized by a UV map $F: \Omega \subset \mathbb{R}^2 \rightarrow \mathbb{R}^3$ mapping 2D coordinates $(u, v) \in \Omega$ to points on the 3D surface.
Beyond its common use in texture mapping, UV-Net~\cite{jayaraman2021uvnet} introduced UV-grids as B-rep face descriptors by rasterizing $\Omega$ with an $n_u \times n_v$ regular lattice (\cref{img:uv_grid}).
This yields a fixed-size grid of surface samples for any type of B-rep surface, each encoding absolute 3D coordinates $(x, y, z)$, surface normals $(n_x, n_y, n_z)$, and a trim mask indicating valid surface regions.
Processing UV-grids with CNNs captures local surface geometry from grid patterns while also encoding global context across faces through absolute positions and orientations.
As shown in \cref{img:global_frame}, UV-grids thus represent per-face point-cloud patches in the global coordinate system, making them highly sensitive to rotations of the CAD model.

\vspace{0.5em}
\noindent \textbf{Local reference frame UV-grids.} 
To decouple UV-grids from global pose and encode pure surface geometry, we express them in per-face Local Reference Frames (LRFs), a common technique in point-cloud learning~\cite{zhang2020global,kim2020rotation}.
The key insight is that local coordinate systems are anchored to point neighborhoods that co-rotate with the geometry, ensuring invariant features within the local frame.
Unlike point clouds, where LRF construction is often challenging due to noise and the need to establish consistent reference directions from point patches~\cite{tombari2010unique, kim2020rotation,zhang2020global}, B-rep faces naturally provide per-face point patch discretization and reference directions through their UV parameterization.
For each face, we construct an orthonormal basis at the face center $\mathbf{o}$, corresponding to the surface point at the UV domain's center.
$\mathbf{N}$ is the outward surface normal in $\mathbf{o}$, $\mathbf{U}$ is obtained by projecting the surface tangent in the parametric $U$-direction onto the tangent plane and normalizing it, and $\mathbf{V} = \mathbf{N} \times \mathbf{U}$.
The resulting frame $\mathbf{R}_f = [\mathbf{U}~\mathbf{V}~\mathbf{N}] \in \mathbf{SO}(3)$ transforms points and normals via:
\begin{equation}
\mathbf{p}' = \mathbf{R}_f^\top (\mathbf{p} - \mathbf{o}), \qquad
\mathbf{n}'_p = \mathbf{R}_f^\top \mathbf{n}_p.
\label{eq:lrf_projection}
\end{equation}
Because both the face center $\mathbf{o}$ and frame $\mathbf{R}_f$ co-rotate with the face under a global rotation, the transformed features $\mathbf{p}'$ and $\mathbf{n}'_p$ remain unchanged.
Consequently, identical faces in different poses (shown by color in \cref{img:lrf_uv_construction}) produce identical LRF UV grids.
The resulting $n_u \times n_v \times 7$ LRF UV tensor encodes rotation-invariant local geometry through relative coordinates, normals, and a trim mask.
We adopt $n_u = n_v = 10$ in all experiments, following UV-Net~\cite{jayaraman2021uvnet}.

\begin{figure}[t]
  \centering
  \begin{subfigure}{0.48\linewidth}
    \centering
    \includegraphics[width=\linewidth]{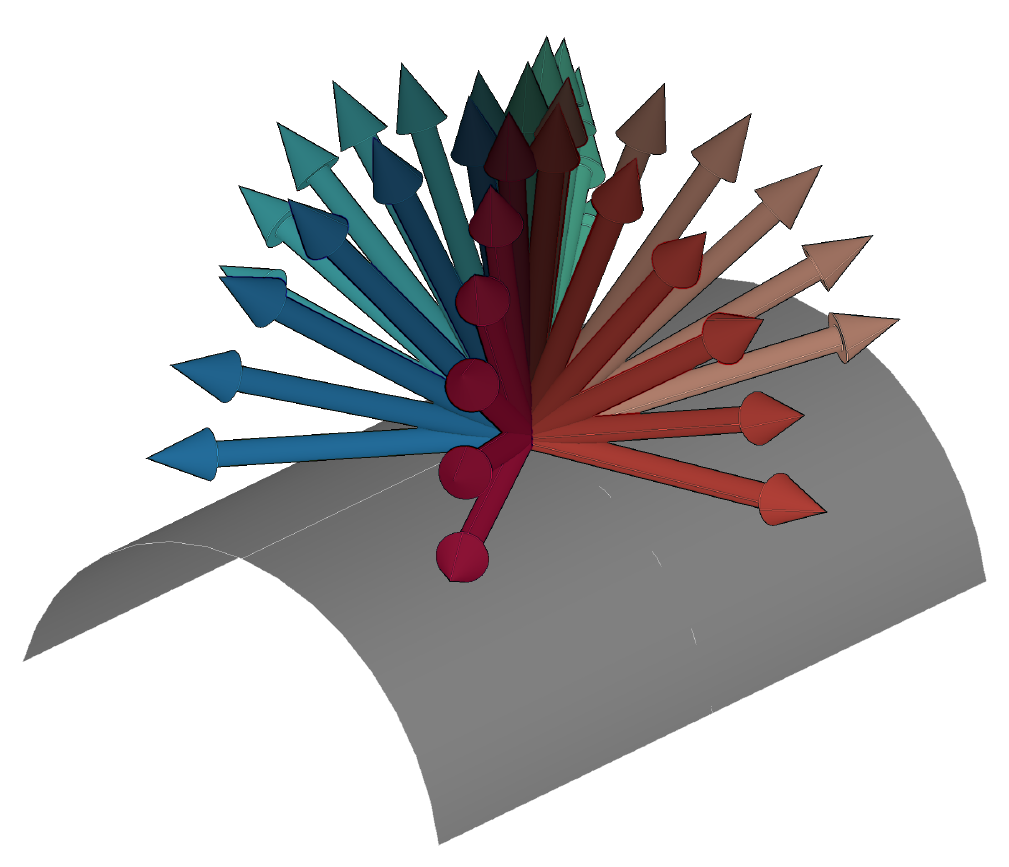}
    \caption{Hemisphere sampling}
    \label{img:hemisphere_a}
  \end{subfigure}
  \hfill
  \begin{subfigure}{0.48\linewidth}
    \centering
    \includegraphics[width=\linewidth]{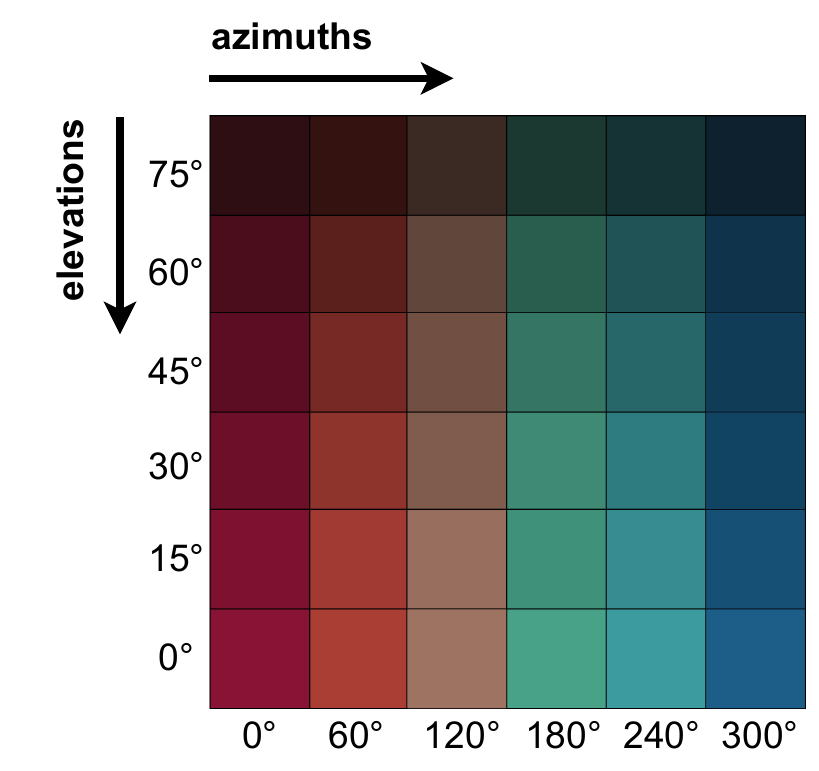}
    \caption{Ray-to-grid mapping}
    \label{img:hemisphere_b}
  \end{subfigure}
  \caption{Field-of-view sampling.
    (a) Rays are cast from the face center $\mathbf{o}$ over a hemisphere oriented along the normal $\mathbf{N}$. 
    Azimuth ($0^\circ$) is aligned with the $\mathbf{U}$ direction.
    (b) The hemisphere is discretized into an elevation $\times$ azimuth grid, forming a 2D descriptor suitable for CNN processing.}\label{img:hemisphere_sampling}
\end{figure}

\begin{figure}[t]
  \centering
  \begin{subfigure}{0.48\linewidth}
    \centering
    \includegraphics[width=\linewidth]{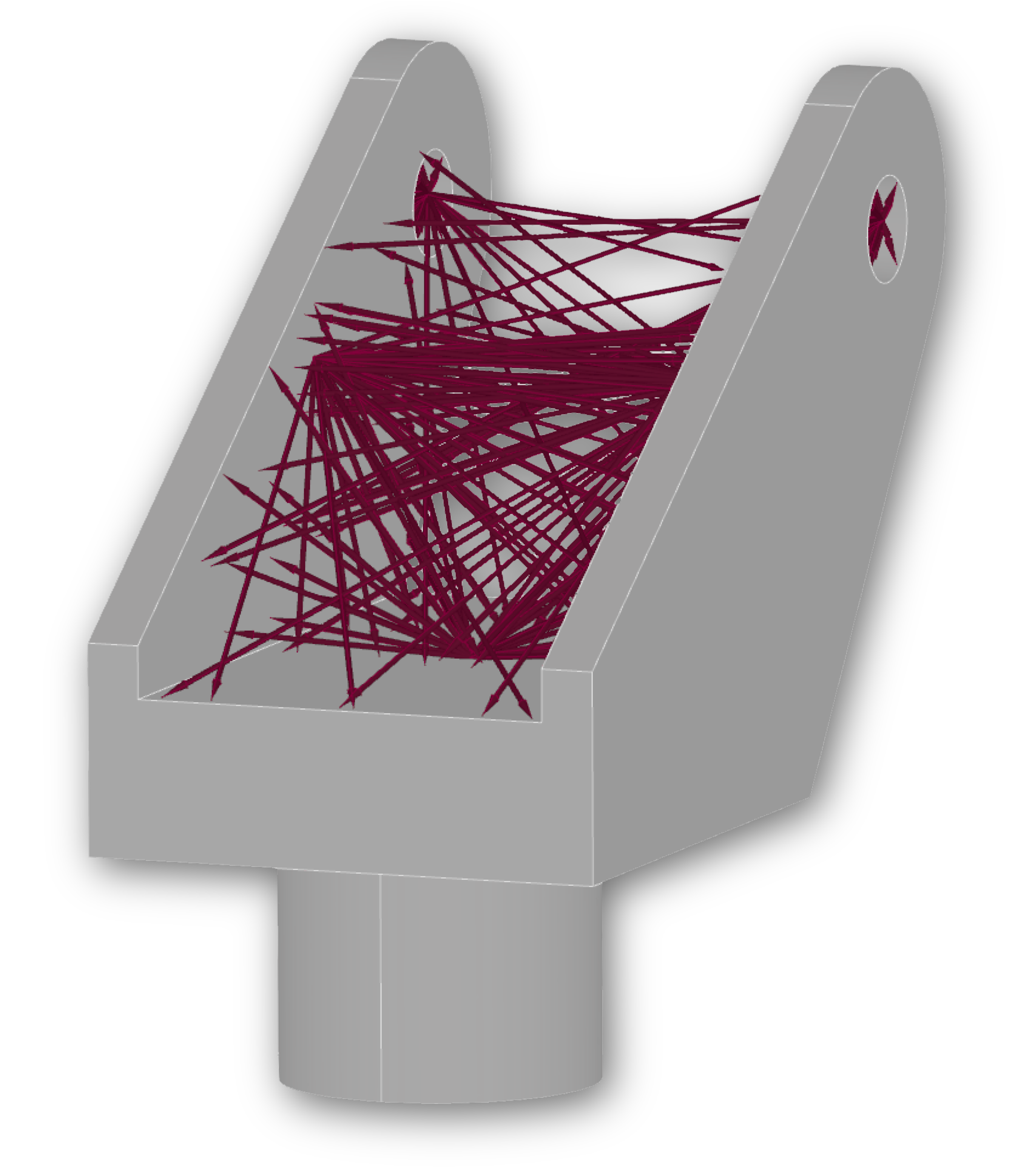}
    \caption{Outward ray casting}
    \label{img:outer_rays}
  \end{subfigure}
  \hfill
  \begin{subfigure}{0.48\linewidth}
    \centering
    \includegraphics[width=\linewidth]{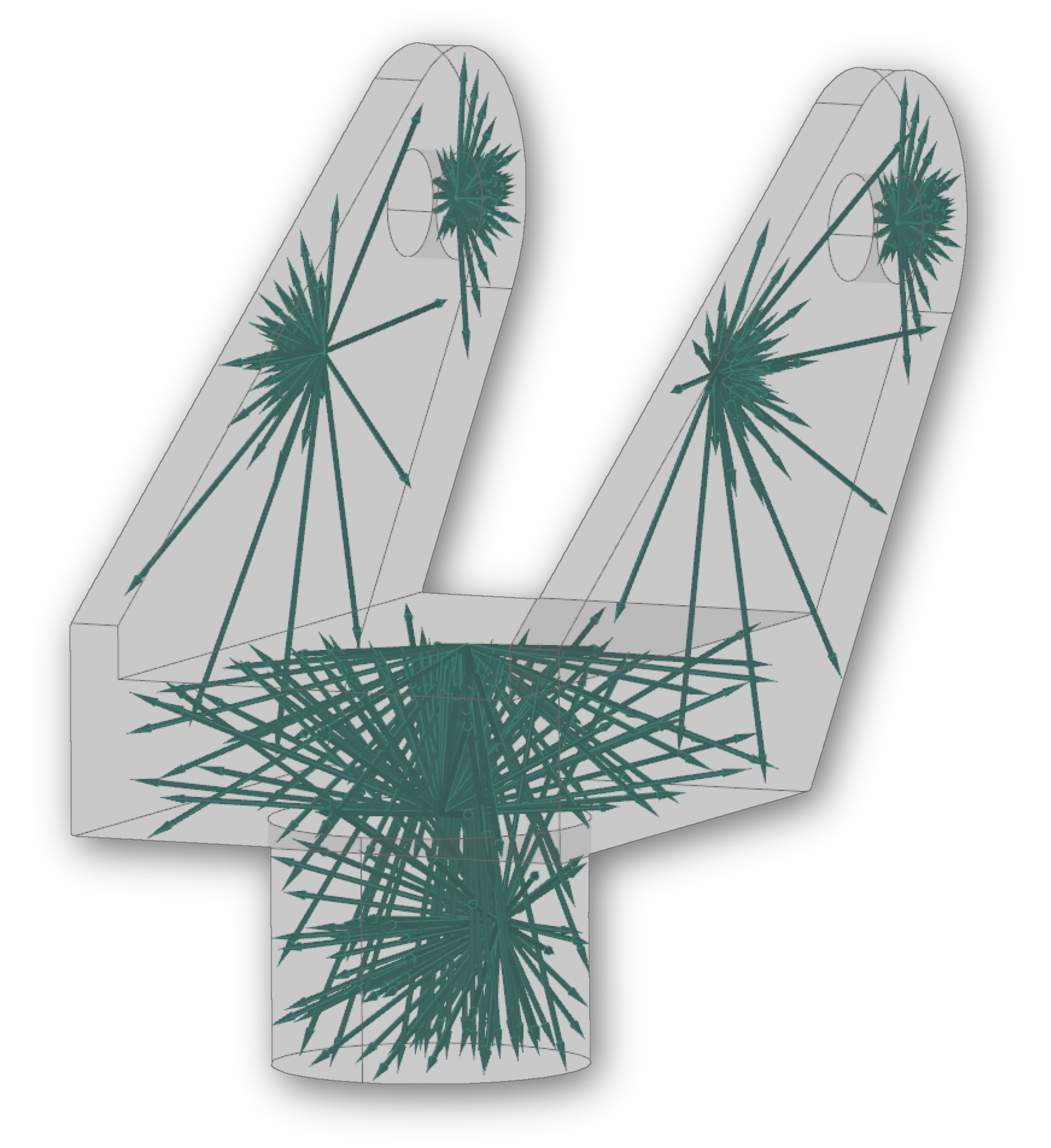}
    \caption{Inward ray casting}
    \label{img:inner_rays}
  \end{subfigure}
  \caption{
    Outward vs.\ inward vision.
    (a) Outward rays (around $\mathbf{N}$) probe the external environment, often resulting in faces that see empty space.
    (b) Inward rays (around $-\mathbf{N}$) probe the interior of the solid, typically yielding dense intersections.
    Only intersection hits displayed.}\label{img:outer_inner_rays}
\end{figure}

\vspace{0.5em}
\noindent \textbf{Field-of-View descriptors.}
While LRF UV grids successfully capture local geometry independent of global pose, they lose structural context in the process.
To recover this information in a rotation-invariant manner, we introduce per-face Field-of-View (FoV) descriptors computed via ray casting, as illustrated in \cref{img:hemisphere_sampling}.
For each face, rays are emitted from its center $\mathbf{o}$ over a hemisphere oriented along the face normal.
The hemisphere is discretized into $n_{\text{el}}$ elevation bins ($0$–$90^\circ$) and $n_{\text{az}}$ azimuth bins ($0$–$360^\circ$), with the azimuth origin ($0^\circ$) aligned to the $\mathbf{U}$ direction and increasing toward $\mathbf{V}$ (\cref{img:hemisphere_a}).
When projected onto a 2D grid, elevation corresponds to rows and azimuth to columns, with each cell representing a unique ray direction (\cref{img:hemisphere_b}).
Because the ray origin and directions are expressed in the local frame $\mathbf{R}_f$, they co-rotate with the face and align with its parameterization, ensuring rotation invariance similar to LRF UV grids.
The resulting Outward Vision (OV) grid captures the external surroundings of each face, providing structural context beyond local geometry as shown in \cref{img:outer_rays}.

Ray intersections are computed using the curve-face intersection functionalities of PythonOCC~\cite{paviot2022pythonocc}, a Python wrapper for the OpenCASCADE CAD kernel.
For each ray, we record three quantities at the first intersection (if any): (1) a binary hit flag, (2) the distance from the ray origin to the intersection, and (3) the dot product between the ray direction and surface normal, encoding the incidence angle.
Rays with no intersection are set to zero across all channels, with the hit flag explicitly marking the absence of geometry.
The structured ray-to-grid mapping thus represents each face's surrounding as a 3-channel field-of-view “image” of size $n_{\text{el}} \times n_{\text{az}}$, suitable for CNN processing.

\vspace{0.5em}
\noindent \textbf{Outward and inward vision.}
For some faces, \eg, on revolving parts, all outward rays may miss geometry, producing sparse or even empty OV-grids.
While the absence of hits already conveys structural cues, we enrich the representation with Inward Vision (IV) grids, generated by casting rays into the opposite hemisphere, around $-\mathbf{N}$ instead of $\mathbf{N}$ (\cref{img:inner_rays}).
The intuition is that outward rays probe the external environment, whereas inward rays capture internal B-rep structure.
In watertight solids, inward rays intersect geometry densely, yielding rich distance information, although watertightness is not required.
In all experiments, we use $n_{\text{el}} = 6$ and $n_{\text{az}} = 12$, corresponding to $15^\circ$ elevation and $30^\circ$ azimuth steps.
The non-square grid naturally reflects the differing ranges of elevation ($90^\circ$) and azimuth ($360^\circ$).
Together, OV and IV grids provide complementary, rotation-invariant representations of external and internal structural context.
Note that our descriptors are also translation-invariant by construction, since they are defined relative to the face center.
While translations are often handled through centering~\cite{jayaraman2021uvnet,wu2024aagnet}, our formulation guarantees this inherently.

\subsection{Network Architecture}
\begin{figure*}
  \centering
  \includegraphics[width=\textwidth]{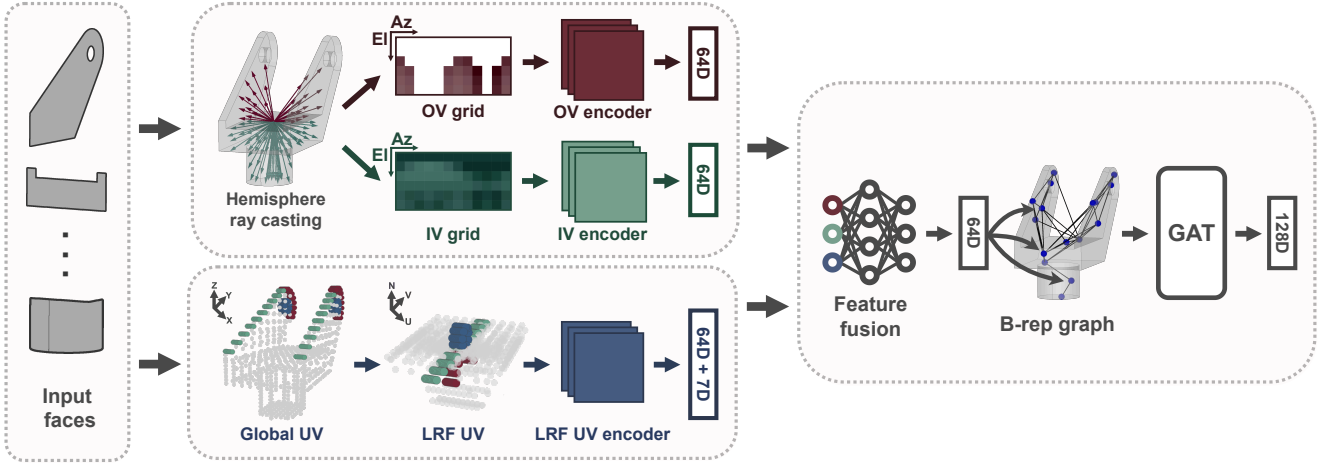}
  \caption{FoV-Net architecture overview.
  For each B-rep face, dedicated CNNs process the Local Reference Frame (LRF) UV grid, Outward Vision (OV) grid, and Inward Vision (IV) grid.
  The resulting feature vectors are concatenated with per-face surface type and area features (7D) and fused by an MLP.
  These serve as node embeddings in the B-rep graph, where a graph attention network (GAT) performs message passing for task-specific learning.
}\label{img:fovnet_architecture}
\end{figure*}

Our Field-of-View Network (FoV-Net) processes the proposed face descriptors with lightweight CNNs, fuses their embeddings, and propagates information across the B-rep graph using a graph attention network (GAT), as shown in \cref{img:fovnet_architecture}.

\vspace{0.5em}
\noindent \textbf{Feature extraction and fusion.}
For each face, FoV-Net takes as input the OV grid ($n_{\text{el}} \times n_{\text{az}} \times 3$), IV grid ($n_{\text{el}} \times n_{\text{az}} \times 3$), and LRF UV grid ($n_u \times n_v \times 7$).
Each tensor is processed by a dedicated lightweight 2D CNN.
The OV and IV encoders each consist of two convolutional layers (32$\rightarrow$64 channels), followed by global average pooling and a linear projection to 64 dimensions.
Circular padding along the azimuthal axis ensures seamless wrap-around, reflecting the periodic topology of azimuthal sampling.
The LRF UV encoder uses three convolutional layers (32$\rightarrow$64$\rightarrow$128 channels), also followed by global average pooling and a linear projection to 64 dimensions.
Additionally, we include per-face one-hot encoded surface type (6D) and face area, forming a 7D attribute vector.
Outputs from all three encoders, together with the face attributes, are concatenated and passed through a two-layer fusion MLP (hidden dimension 256) to capture cross-view interactions, yielding a unified 64-D per-face node embedding.
While prior work has included edge-level U-grids and features, we omit them, as ablation studies have shown that their inclusion yields minimal performance gains~\cite{jayaraman2021uvnet}, while introducing additional computational overhead due to per-edge forward passes.

\vspace{0.5em}
\noindent \textbf{Graph attention network backbone.}
Given the face adjacency graph $\mathcal{G}$ and per-face node embeddings, a 3-layer graph attention network (GAT)~\cite{velivckovic2017graph} with 4 attention heads and hidden dimension 64 propagates information across neighboring faces: 
\[
\mathbf{h}_i^{(\ell+1)} = \sigma\Bigg(\sum_{j \in \mathcal{N}(i)} \alpha_{ij}^{(\ell)} \mathbf{W}^{(\ell)} \mathbf{h}_j^{(\ell)}\Bigg),
\]
where $\mathbf{h}_i^{(\ell)}$ is the hidden representation of face $i$ at layer $\ell$, $\mathcal{N}(i)$ its neighboring faces, $\mathbf{W}^{(\ell)}$ a learnable weight matrix, $\alpha_{ij}^{(\ell)}$ the attention coefficient, and $\sigma$ a ReLU nonlinearity. 
For classification, node embeddings are globally aggregated via max pooling and fed into a two-layer MLP classifier. 
For segmentation, the final node embeddings are passed directly to a per-face prediction head.
\section{Experiments}
\label{sec:experiments}

We evaluate FoV-Net against state-of-the-art B-rep learning methods on multiple classification and segmentation benchmarks.
Our experiments are designed to: 
(1) assess performance under canonical and randomly rotated orientations, 
(2) ablate the contributions of LRF UV and FoV features and vision-grid resolution, and 
(3) evaluate data efficiency under limited supervision.

\subsection{Datasets}

\noindent \textbf{SolidLetters}~\cite{jayaraman2021uvnet} contains 96k shapes generated by extruding and filleting upper- and lower-case alphabet letters across 2002 fonts.  
The task is 26-way letter classification.  
Strong intra-class variation (case and font) and high inter-class similarity (\eg, O vs.\ Q) make this a challenging benchmark for geometric understanding.

\vspace{0.5em}
\noindent \textbf{TraceParts}~\cite{mandelli2022cad} contains 600 mechanical parts across six categories (screws, nuts, hinges, fans, iron bars, wheels).
This small, simple dataset exhibits large inter-class variation and serves as a sanity check for rotation sensitivity.

\vspace{0.5em}
\noindent \textbf{Fusion360 Gallery}~\cite{lambourne2021brepnet} comprises 35,858 CAD models from Autodesk Fusion 360 users, annotated with per-face segmentation labels corresponding to modeling operations (8 classes, \eg, ExtrudeSide, CutEnd, Fillet).

\vspace{0.5em}
\noindent \textbf{MFCAD++}~\cite{colligan2022hierarchical} contains 59,655 cuboid-based CAD models annotated with 25 machining features (\eg, pocket, hole, chamfer).
Each model has 3--10 labeled features used for per-face segmentation.

\subsection{Models}
We compare FoV-Net with two leading B-rep learning methods: UV-Net~\cite{jayaraman2021uvnet} and AAGNet~\cite{wu2024aagnet}.  

\vspace{0.5em}
\noindent \textbf{UV-Net} samples points and normals from each face's UV domain to construct per-face and per-edge embeddings using lightweight CNNs followed by a GNN backbone.  
Relying solely on UV features, it directly tests the rotation sensitivity of UV-based descriptors.  
We adopt the original configuration with $n_u = n_v = 10$.

\vspace{0.5em}
\noindent \textbf{AAGNet} extends UV-based features with explicit geometric descriptors for faces (\eg, surface type, area) and edges (\eg, edge type, length, convexity).  
These local features could provide additional rotation-invariant information.  
We adopt the original architecture with $n_u = n_v = 10$.

\vspace{0.5em}
\noindent \textbf{UV-Net (aug)} and \textbf{AAGNet (aug)} denote rotation-augmented variants, where each training sample is randomly rotated in $\mathbf{SO}(3)$ at every access during training. 
This setting tests whether data augmentation can alleviate rotation sensitivity in UV-based methods.

\vspace{0.5em}
\noindent \textbf{FoV-Net (UV)} uses the FoV-Net architecture with global UV grids instead of LRF UV grids.
This variant is not rotation-invariant by design; it assesses whether adding FoV descriptors improves performance on aligned datasets when combined with traditional, rotation-sensitive UV features.

\subsection{Experimental setup}
Models are evaluated on both original and randomly rotated test sets, with rotations sampled from $\mathbf{SO}(3)$ around the model center using axis-angle representations.
Validation sets remain unrotated to reflect original performance during model selection. 
All CAD models are normalized to a unit cube centered at the origin using the bounding box diagonal~\cite{jayaraman2021uvnet,wu2024aagnet}. 
We use the official train/val/test splits for Fusion360, MFCAD++, and SolidLetters, and a stratified 80/10/10 split for TraceParts.
Training uses Adam (lr=0.001, batch size 64) with early stopping (patience 30).
All experiments were conducted on a single NVIDIA RTX A5000 (24~GB VRAM).  
Metrics are multiclass accuracy for classification, and per-face accuracy and mean Intersection over Union (IoU) for segmentation.
Each experiment is repeated five times with different random seeds, reporting mean~$\pm$~standard deviation.

\subsection{Results}

\noindent \textbf{Classification.} 
\Cref{tab:classification_results} reports accuracy on SolidLetters and TraceParts.
FoV-Net maintains identical performance under rotations, whereas UV-Net and AAGNet suffer severe drops, revealing strong dependence on absolute pose.
On SolidLetters, UV-Net loses 87.41 percentage points (pp) and AAGNet 82.65 pp under rotation.
On the smaller TraceParts dataset, all models reach perfect original accuracy, but UV-Net still drops 54.33 pp and AAGNet 8.67 pp, the latter likely benefiting from rotation-invariant attributes (\eg, surface type, area).
Rotation augmentation substantially improves rotation-invariance, yet performance remains below FoV-Net and drops relative to their own unaugmented counterparts.
On original SolidLetters, UV-Net and AAGNet can exploit some pose cues (\eg, b vs. d, p vs. q), while FoV-Net—being rotation-invariant—cannot.
Despite this, FoV-Net remains competitive, demonstrating the strength of learning intrinsic geometry and structural context.
Adding global UV features to FoV-Net (FoV-Net (UV)) boosts original accuracy to 97.42\%, showing that FoV features can complement UV cues when absolute pose is informative, but at the cost of rotation-invariance.
\begin{table}[t]
    \small
    \centering
    \caption{Classification accuracy (\%) on SolidLetters and TraceParts datasets.
    Mean ± std results over five runs are reported for rotated and original test sets, with best results in bold.}
    \begin{tabular}{p{1.3cm} p{1.9cm} c c}
        \toprule
        & & \multicolumn{2}{c}{Accuracy}\\
		\cmidrule(lr){3-4}
        Dataset & Model & Rotated & Original\\
        \midrule
        \multirow{6}{*}{SolidLetters} 
        & FoV-Net       & \textbf{96.35 ± 0.16}     & 96.35 ± 0.16 \\
        & FoV-Net (UV)  & 23.53 ± 0.61              & \textbf{97.42 ± 0.17}\\
        & UV-Net        & 8.94 ± 0.46               & 97.10 ± 0.15\\
        & AAGNet        & 14.03 ± 0.51              & 96.68 ± 0.19\\        
        & UV-Net (aug)  & 94.97 ± 0.10              & 95.34 ± 0.12\\
        & AAGNet (aug)  & 94.43 ± 0.22              & 94.95 ± 0.18\\
        \midrule
        \multirow{6}{*}{TraceParts} 
        & FoV-Net       & \textbf{100.00 ± 0.00}    & \textbf{100.00 ± 0.00} \\
        & FoV-Net (UV)  & 93.33 ± 9.20              & \textbf{100.00 ± 0.00}\\
        & UV-Net        & 45.67 ± 6.19              & \textbf{100.00 ± 0.00} \\
        & AAGNet        & 91.33 ± 4.92              & \textbf{100.00 ± 0.00} \\        
        & UV-Net (aug)  & 98.33 ± 2.36              & \textbf{100.00 ± 0.00} \\
        & AAGNet (aug)  & \textbf{100.00 ± 0.00}    & \textbf{100.00 ± 0.00}\\        
        \bottomrule
    \end{tabular}
    \label{tab:classification_results}
\end{table}

\vspace{0.5em}
\noindent \textbf{Segmentation.} 
\Cref{tab:segmentation_results} reports segmentation accuracy and IoU on Fusion360 and MFCAD++.
\begin{table*}[t]
    \small
    \centering
    \caption{Segmentation accuracy (\%) and Intersection over Union (\%) on Fusion360 and MFCAD++ datasets. 
    Mean ± std results over five runs are reported for rotated and original test sets, with best results in bold.}
    \begin{tabular}{
        p{0.15\textwidth} 
        p{0.15\textwidth} 
        >{\centering}p{0.14\textwidth} 
        >{\centering}p{0.14\textwidth} 
        >{\centering}p{0.14\textwidth}
        >{\centering\arraybackslash}p{0.14\textwidth}  
        }      
        \toprule
        & & \multicolumn{2}{c}{Accuracy} & \multicolumn{2}{c}{IoU}\\
        \cmidrule(lr){3-4} \cmidrule(lr){5-6}
        Dataset & Model & Rotated & Original & Rotated & Original \\
        \midrule
        \multirow{6}{*}{Fusion360} 
        & FoV-Net       & \textbf{91.72 ± 0.17} & \textbf{91.72 ± 0.17} & \textbf{73.81 ± 0.62} & \textbf{73.81 ± 0.62}\\
        & FoV-Net (UV)  & 81.37 ± 0.32          & 89.46 ± 0.48          & 54.89 ± 0.66          & 69.71 ± 1.34 \\
        & UV-Net        & 69.13 ± 2.18          & 88.34 ± 0.41          & 37.07 ± 1.26          & 66.48 ± 0.56 \\
        & AAGNet        & 79.85 ± 0.92          & 90.60 ± 0.25          & 53.42 ± 0.84          & \textbf{73.81 ± 0.74} \\  
        & UV-Net (aug)  & 84.26 ± 1.18          & 86.04 ± 0.88          & 59.90 ± 2.27          & 63.16 ± 2.11 \\
        & AAGNet (aug)  & 86.70 ± 0.28          & 87.91 ± 0.56          & 66.40 ± 0.72          & 68.49 ± 1.07 \\        
        \midrule
        \multirow{6}{*}{MFCAD++} 
        & FoV-Net       & \textbf{99.33 ± 0.07} & \textbf{99.33 ± 0.07} & \textbf{97.81 ± 0.24} & 97.81 ± 0.24 \\
        & FoV-Net (UV)  & 70.45 ± 3.20          & 99.19 ± 0.06          & 58.87 ± 5.87          & 97.40 ± 0.18 \\
        & UV-Net        & 35.44 ± 3.84          & 98.95 ± 0.06          & 18.79 ± 2.84          & 96.65 ± 0.15 \\
        & AAGNet        & 80.13 ± 1.71          & \textbf{99.33 ± 0.14} & 64.70 ± 1.31          & \textbf{98.10 ± 0.36} \\
        & UV-Net (aug)  & 98.74 ± 0.10          & 98.77 ± 0.11          & 95.86 ± 0.28          & 95.96 ± 0.31 \\
        & AAGNet (aug)  & 99.12 ± 0.19          & 99.12 ± 0.18          & 97.48 ± 0.52          & 97.48 ± 0.52 \\
        \bottomrule
    \end{tabular}
    \label{tab:segmentation_results}
\end{table*}
Trends mirror those in classification, although performance drops under rotation are less severe.
This is likely because, in segmentation, faces occupy varying absolute positions even when the model is aligned, forcing UV-based methods to partially ignore absolute pose cues.
On Fusion360, UV-Net and AAGNet still lose 19.21 and 10.75 percentage points in accuracy under rotation, corresponding to 29.41 and 20.39 pp drops in IoU.
On MFCAD++, UV-Net's accuracy collapses by 63.51 pp (77.86 pp IoU), while AAGNet drops 19.20 pp (33.40 pp IoU).
FoV-Net maintains strong, consistent performance across orientations, only matched or outperformed by AAGNet on aligned test sets.
Rotation augmentation again improves rotation robustness at the cost of reduced overall performance
Unlike SolidLetters, where absolute pose was informative for the original data, FoV-Net outperforms FoV-Net (UV) on both segmentation tasks, suggesting that global UV features do not add useful positional information that can be leveraged here.

\subsection{Ablation studies}
\begin{figure}[t]
    \centering
    \includegraphics[width=\linewidth]{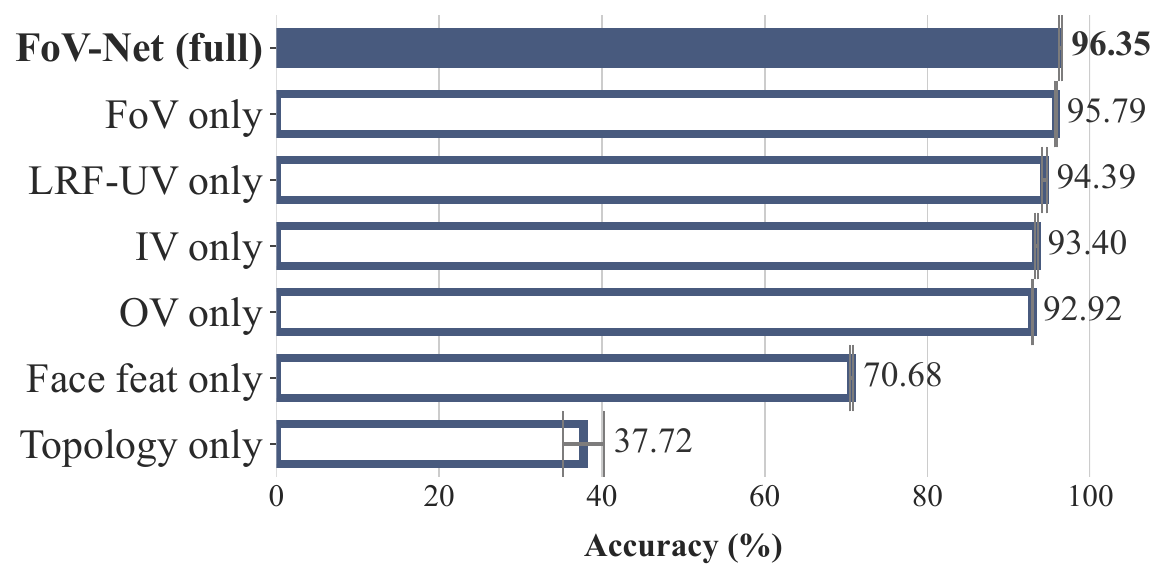}
    \caption{FoV-Net feature ablation on SolidLetters.
    FoV-Net is trained using only one feature type at a time to assess their individual contributions.
    Mean accuracy and standard deviation over five runs are shown.}
    \label{img:feature_ablation}
\end{figure}

We ablate the contribution of FoV-Net's individual feature components on SolidLetters by training variants with only a single feature type: FoV grids, LRF UV grids, OV grids, IV grids, per-face features (six surface-type indicators + area), or topology alone (node degree as the sole input to the GAT).
\cref{img:feature_ablation} shows the mean test accuracy and standard deviation over five runs.
FoV grids (95.79\%) and LRF UV alone (94.39\%) both capture highly informative cues, with their stronger combined performance (96.35\%) demonstrating complementarity.
Using only OV and IV vision grids achieve 92.92\% and 93.40\%, respectively, indicating that both outward and inward visibility provide useful context, with their combination outperforming either alone (95.79\%).
Using only per-face features yields 70.68\%, and topology alone drops to 37.72\%, confirming that our proposed descriptors are essential beyond simple local attributes or graph structure.

\begin{figure}[t]
    \centering
    \includegraphics[width=\linewidth]{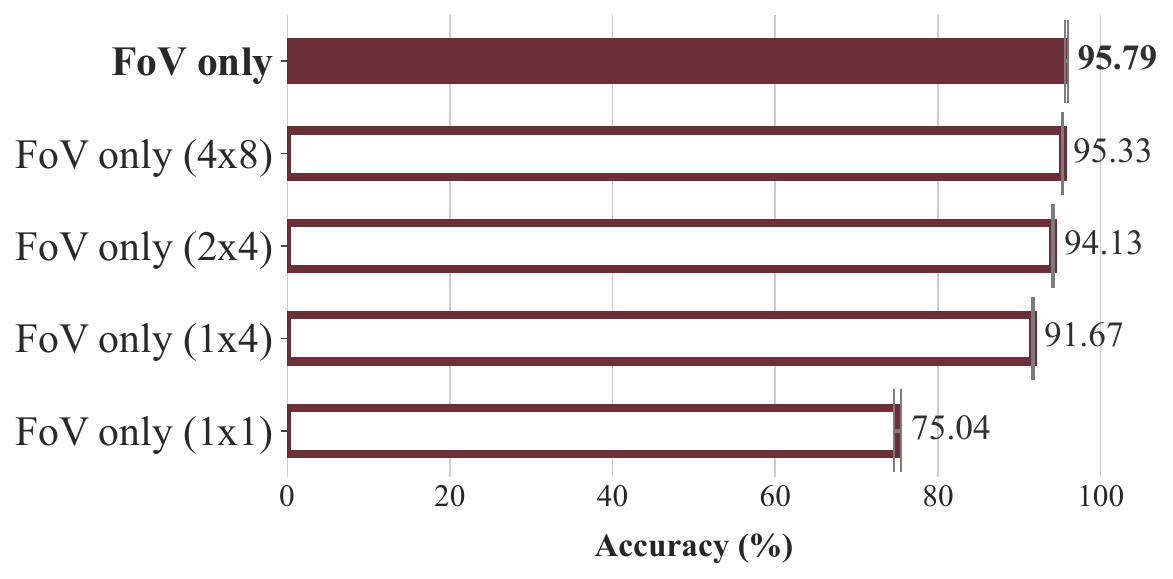}
    \caption{FoV resolution ablation on SolidLetters. 
    Using only FoV features, we vary the vision grid resolution from the original $6\times12$ down to $1\times1$ (single ray along the face normal).
    Mean accuracy and standard deviation over five runs are shown.}
    \label{img:resolution_ablation}
\end{figure}

We study the impact of FoV grid resolution on SolidLetters when using only FoV grids.
The number of elevation and azimuth bins ($n_{\text{el}} \times n_{\text{az}}$) is reduced from the original $6 \times 12$ to $4 \times 8$, $2 \times 4$, $1 \times 4$, and $1 \times 1$, with the last case representing a single ray along the face normal from the face center.
As shown in \cref{img:resolution_ablation}, accuracy declines gradually as the resolution decreases, with only minor losses down to $4 \times 2$.
Performance drops more noticeably at $4 \times 1$ and collapses to 75.04\% at $1 \times 1$, however, highlighting the importance of multi-directional ray sampling over a single ray.

\subsection{Data efficiency analysis}
\begin{figure}
    \centering
    \includegraphics[width=\linewidth]{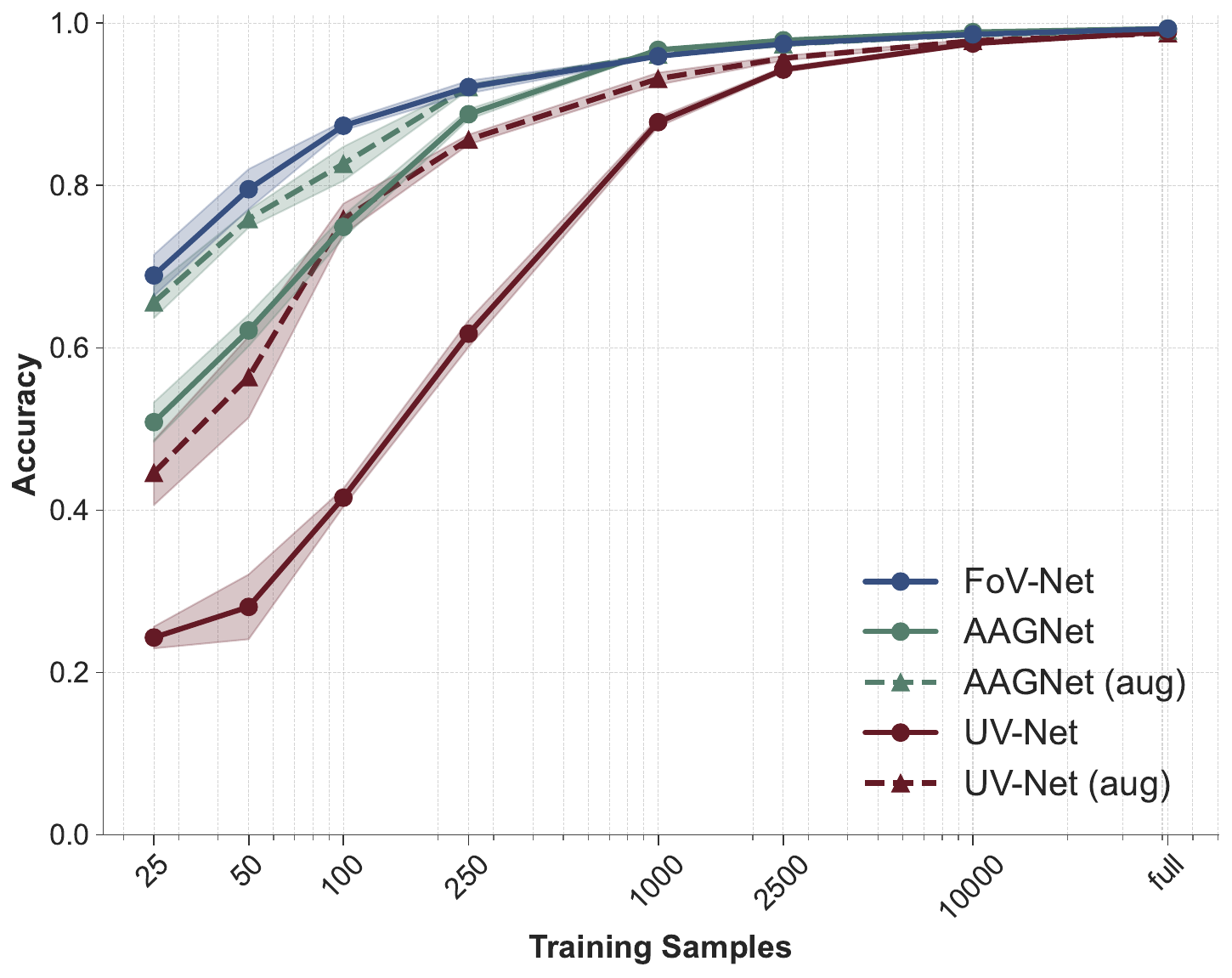}
    \caption{Learning curves on MFCAD++. 
    Training set sizes range from 25 samples to the full dataset (log scale). 
    Mean accuracy and standard deviation over five random subsets are shown.}
    \label{img:learning_curves}
\end{figure}

In segmentation tasks such as feature recognition, target faces often appear at varying absolute positions.
Because UV-based descriptors rely on absolute coordinates and normals, they may inadvertently overfit spurious correlations between face positions and labels even on well-aligned CAD models, especially in low-data regimes.
In contrast, FoV-Net's rotation invariance should improve data efficiency by learning more generalizable intrinsic geometry.
To quantify this, we evaluate data efficiency on the MFCAD++ segmentation dataset, focusing on generalization under limited supervision.
Models are trained on progressively larger random subsets of the training data (25 samples up to the full dataset), with each experiment repeated five times.
The same random subsets are shared across models for fair comparison, and model selection and evaluation are performed on the full validation and test sets.
\cref{img:learning_curves} shows learning curves with standard deviation bands.

FoV-Net achieves 80\% accuracy with just 50 samples, compared to $\sim$60\% for AAGNet and $\sim$30\% for UV-Net. 
AAGNet catches up to FoV-Net at 1,000 samples, while UV-Net requires 10,000 samples, underscoring the poor generalization of rotation-sensitive UV features in low-data regimes.
Edge and surface descriptors give AAGNet an advantage over UV-Net, but it still falls short of FoV-Net in low-data performance.
Rotation augmentation significantly benefits UV-based models on small subsets by disrupting pose correlations, but, as segmentation results in \cref{tab:segmentation_results} have shown, it comes at the cost of overall reduced full-dataset performance.
FoV-Net avoids this trade-off, achieving strong, consistent accuracy across all dataset sizes without augmentation—an important advantage for industrial CAD applications, where data availability is often limited by IP restrictions~\cite{stjepandic2015intellectual}.
These results demonstrate that intrinsic rotation invariance is critical not only for robustness to arbitrary orientations but also for segmentation on well-aligned models, particularly in low-data regimes.
\section{Conclusion}
\label{sec:conclusion}
This work quantifies the rotation sensitivity of existing CAD B-rep learning methods, revealing severe performance degradation under arbitrary orientations.
To address this, we propose FoV-Net, a rotation-invariant framework that combines Local Reference Frame (LRF) UV grids for local geometry with ray-casted Field-of-View (FoV) descriptors for structural context.
FoV-Net achieves strong performance on both classification and segmentation benchmarks, particularly under arbitrary rotations.
We further show that rotation invariance is critical even for well-aligned data in segmentation tasks, as face positions vary across parts and existing methods tend to overfit spurious pose correlations in low-data regimes.
While rotation augmentation partially mitigates low-data issues, augmented models show reduced full-dataset performance.
FoV-Net's inherently rotation-invariant features avoid this trade-off by design.
Overall, our approach represents an important step toward robust, generalizable CAD B-rep learning, with promising implications for manufacturing and industrial applications.

\vspace{0.5em}
\noindent \textbf{Limitations and future work.}
Our ray-casting methodology, built on the PythonOCC~\cite{paviot2022pythonocc} CAD kernel, supports CPU-level parallelization, and even very small FoV grids yield informative features.
Current experiments focus on single-part B-reps of simple to moderate complexity; future work will explore scalability to larger assemblies and FoV grids, potentially leveraging GPU acceleration or mesh-level ray casting.
FoV grids rely on an equiangular 3D-to-2D mapping that introduces polar distortions similar to geographic projections; spherical CNNs~\cite{esteves2018learning,cohen2018spherical} could provide more uniform directional parameterizations.
While FoV-Net ensures rotation invariance, surface reparameterizations (\eg, UV-axis flips or swaps) remain challenging, though UV-Net~\cite{jayaraman2021uvnet} showed promising mitigation via D\textsubscript{2}-equivariant convolutions~\cite{weiler2019general}.  
Beyond invariance, FoV descriptors capture structural context that current benchmarks only weakly emphasize; future datasets targeting more structural tasks like milling tool reachability could better highlight these strengths. 
Our architecture omits edge features for computational reasons, but extensions to B-rep generation where edge features are essential is an important direction.
Finally, our descriptors demonstrate strong generalization, providing a foundation for contrastive B-rep pretraining and unsupervised CAD retrieval.

{
    \small
    \bibliographystyle{ieeenat_fullname}
    \bibliography{main}

@String(CVPR= {IEEE Conf. Comput. Vis. Pattern Recog.})

@String(ECCV= {Eur. Conf. Comput. Vis.})

@String(TOG= {ACM Trans. Graph.})

@String(CVPR  = {CVPR})

@String(ECCV  = {ECCV})

@String(TOG   = {ACM TOG})

@article{ansaldi1985geometric,
  title     = {Geometric modeling of solid objects by using a face adjacency graph representation},
  author    = {Ansaldi, Silvia and De Floriani, Leila and Falcidieno, Bianca},
  journal   = {ACM SIGGRAPH Computer Graphics},
  volume    = {19},
  number    = {3},
  pages     = {131--139},
  year      = {1985},
  publisher = {ACM New York, NY, USA}
}

@inproceedings{bai2020d3feat,
  title     = {D3feat: Joint learning of dense detection and description of 3d local features},
  author    = {Bai, Xuyang and Luo, Zixin and Zhou, Lei and Fu, Hongbo and Quan, Long and Tai, Chiew-Lan},
  booktitle = {Proceedings of the IEEE/CVF conference on computer vision and pattern recognition},
  pages     = {6359--6367},
  year      = {2020}
}

@inproceedings{cao2020graph,
  title        = {Graph representation of 3D CAD models for machining feature recognition with deep learning},
  author       = {Cao, Weijuan and Robinson, Trevor and Hua, Yang and Boussuge, Flavien and Colligan, Andrew R and Pan, Wanbin},
  booktitle    = {International design engineering technical conferences and computers and information in engineering conference},
  volume       = {84003},
  pages        = {V11AT11A003},
  year         = {2020},
  organization = {American Society of Mechanical Engineers}
}

@inproceedings{chen2019clusternet,
  title     = {Clusternet: Deep hierarchical cluster network with rigorously rotation-invariant representation for point cloud analysis},
  author    = {Chen, Chao and Li, Guanbin and Xu, Ruijia and Chen, Tianshui and Wang, Meng and Lin, Liang},
  booktitle = {Proceedings of the IEEE/CVF conference on computer vision and pattern recognition},
  pages     = {4994--5002},
  year      = {2019}
}

@inproceedings{chen2024local,
  title     = {Local-consistent transformation learning for rotation-invariant point cloud analysis},
  author    = {Chen, Yiyang and Duan, Lunhao and Zhao, Shanshan and Ding, Changxing and Tao, Dacheng},
  booktitle = {Proceedings of the IEEE/CVF Conference on Computer Vision and Pattern Recognition},
  pages     = {5418--5427},
  year      = {2024}
}

@inproceedings{choy2019fully,
  title     = {Fully convolutional geometric features},
  author    = {Choy, Christopher and Park, Jaesik and Koltun, Vladlen},
  booktitle = {Proceedings of the IEEE/CVF international conference on computer vision},
  pages     = {8958--8966},
  year      = {2019}
}

@article{cohen2018spherical,
  title   = {Spherical cnns},
  author  = {Cohen, Taco S and Geiger, Mario and K{\"o}hler, Jonas and Welling, Max},
  journal = {arXiv preprint arXiv:1801.10130},
  year    = {2018}
}

@article{colligan2022hierarchical,
  title     = {Hierarchical cadnet: Learning from b-reps for machining feature recognition},
  author    = {Colligan, Andrew R and Robinson, Trevor T and Nolan, Declan C and Hua, Yang and Cao, Weijuan},
  journal   = {Computer-Aided Design},
  volume    = {147},
  pages     = {103226},
  year      = {2022},
  publisher = {Elsevier}
}

@inproceedings{deng2018ppf,
  title     = {Ppf-foldnet: Unsupervised learning of rotation invariant 3d local descriptors},
  author    = {Deng, Haowen and Birdal, Tolga and Ilic, Slobodan},
  booktitle = {Proceedings of the European conference on computer vision (ECCV)},
  pages     = {602--618},
  year      = {2018}
}

@inproceedings{deng2021vector,
  title     = {Vector neurons: A general framework for so (3)-equivariant networks},
  author    = {Deng, Congyue and Litany, Or and Duan, Yueqi and Poulenard, Adrien and Tagliasacchi, Andrea and Guibas, Leonidas J},
  booktitle = {Proceedings of the IEEE/CVF international conference on computer vision},
  pages     = {12200--12209},
  year      = {2021}
}

@inproceedings{esteves2018learning,
  title     = {Learning so (3) equivariant representations with spherical cnns},
  author    = {Esteves, Carlos and Allen-Blanchette, Christine and Makadia, Ameesh and Daniilidis, Kostas},
  booktitle = {Proceedings of the european conference on computer vision (ECCV)},
  pages     = {52--68},
  year      = {2018}
}

@article{fan2024neuronurbs,
  title    = {NeuroNURBS: Learning Efficient Surface Representations for 3D Solids},
  author   = {Fan, Jiajie and Gholami, Babak and B{\"a}ck, Thomas and Wang, Hao},
  journal  = {arXiv preprint arXiv:2411.10848},
  year     = {2024}
}

@article{fei2024rotation,
  title     = {Rotation invariance and equivariance in 3D deep learning: a survey},
  author    = {Fei, Jiajun and Deng, Zhidong},
  journal   = {Artificial Intelligence Review},
  volume    = {57},
  number    = {7},
  pages     = {168},
  year      = {2024},
  publisher = {Springer}
}

@inproceedings{fu2025consistent,
  title     = {Consistent Normal Orientation for 3D Point Clouds via Least Squares on Delaunay Graph},
  author    = {Fu, Rao and Zheng, Jianmin and Yu, Liang},
  booktitle = {Proceedings of the Computer Vision and Pattern Recognition Conference},
  pages     = {16932--16942},
  year      = {2025}
}

@article{fuchs2020se,
  title   = {Se (3)-transformers: 3d roto-translation equivariant attention networks},
  author  = {Fuchs, Fabian and Worrall, Daniel and Fischer, Volker and Welling, Max},
  journal = {Advances in neural information processing systems},
  volume  = {33},
  pages   = {1970--1981},
  year    = {2020}
}

@article{guo2022complexgen,
  title     = {Complexgen: Cad reconstruction by b-rep chain complex generation},
  author    = {Guo, Haoxiang and Liu, Shilin and Pan, Hao and Liu, Yang and Tong, Xin and Guo, Baining},
  journal   = {ACM Transactions on Graphics (TOG)},
  volume    = {41},
  number    = {4},
  pages     = {1--18},
  year      = {2022},
  publisher = {ACM New York, NY, USA}
}

@inproceedings{guo2025brepgiff,
  title     = {BrepGiff: Lightweight Generation of Complex B-rep with 3D GAT Diffusion},
  author    = {Guo, Hao and Huang, Xiaoshui and Bai, Yunpeng and Gan, Hongping and Shi, Yilei and others},
  booktitle = {Proceedings of the Computer Vision and Pattern Recognition Conference},
  pages     = {26587--26596},
  year      = {2025}
}

@article{hanocka2019meshcnn,
  title     = {Meshcnn: a network with an edge},
  author    = {Hanocka, Rana and Hertz, Amir and Fish, Noa and Giryes, Raja and Fleishman, Shachar and Cohen-Or, Daniel},
  journal   = {ACM Transactions on Graphics (ToG)},
  volume    = {38},
  number    = {4},
  pages     = {1--12},
  year      = {2019},
  publisher = {ACM New York, NY, USA}
}

@article{heidari2025geometric,
  title     = {Geometric deep learning for computer-aided design: A survey},
  author    = {Heidari, Negar and Iosifidis, Alexandros},
  journal   = {IEEE Access},
  year      = {2025},
  publisher = {IEEE}
}

@inproceedings{hu2020you,
  title     = {What you see is what you get: Exploiting visibility for 3d object detection},
  author    = {Hu, Peiyun and Ziglar, Jason and Held, David and Ramanan, Deva},
  booktitle = {Proceedings of the IEEE/CVF Conference on Computer Vision and Pattern Recognition},
  pages     = {11001--11009},
  year      = {2020}
}

@inproceedings{jayaraman2021uvnet,
  author    = {Jayaraman, Pradeep Kumar and Sanghi, Aditya and Lambourne, Joseph G and Willis, Karl DD and Davies, Thomas and Shayani, Hooman and Morris, Nigel},
  booktitle = {Proceedings of the IEEE/CVF Conference on Computer Vision and Pattern Recognition (CVPR)},
  title     = {Uv-net: Learning from boundary representations},
  pages     = {11703--11712},
  year      = {2021}
}

@inproceedings{jones2023self,
  title     = {Self-supervised representation learning for cad},
  author    = {Jones, Benjamin T and Hu, Michael and Kodnongbua, Milin and Kim, Vladimir G and Schulz, Adriana},
  booktitle = {Proceedings of the IEEE/CVF conference on computer vision and pattern recognition},
  pages     = {21327--21336},
  year      = {2023}
}

@article{joshi1988graph,
  title     = {Graph-based heuristics for recognition of machined features from a 3D solid model},
  author    = {Joshi, Sanjay and Chang, Tien-Chien},
  journal   = {Computer-aided design},
  volume    = {20},
  number    = {2},
  pages     = {58--66},
  year      = {1988},
  publisher = {Elsevier}
}

@article{kim2020rotation,
  title   = {Rotation-invariant local-to-global representation learning for 3d point cloud},
  author  = {Kim, Seohyun and Park, Jaeyoo and Han, Bohyung},
  journal = {Advances in Neural Information Processing Systems},
  volume  = {33},
  pages   = {8174--8185},
  year    = {2020}
}

@inproceedings{koch2019abc,
  title     = {Abc: A big cad model dataset for geometric deep learning},
  author    = {Koch, Sebastian and Matveev, Albert and Jiang, Zhongshi and Williams, Francis and Artemov, Alexey and Burnaev, Evgeny and Alexa, Marc and Zorin, Denis and Panozzo, Daniele},
  booktitle = {Proceedings of the IEEE/CVF conference on computer vision and pattern recognition},
  pages     = {9601--9611},
  year      = {2019}
}

@inproceedings{lambourne2021brepnet,
  title     = {Brepnet: A topological message passing system for solid models},
  author    = {Lambourne, Joseph G and Willis, Karl DD and Jayaraman, Pradeep Kumar and Sanghi, Aditya and Meltzer, Peter and Shayani, Hooman},
  booktitle = {Proceedings of the IEEE/CVF conference on computer vision and pattern recognition},
  pages     = {12773--12782},
  year      = {2021}
}

@inproceedings{lee2001partial,
  title     = {Partial entity structure: a compact non-manifold boundary representation based on partial topological entities},
  author    = {Lee, Sang Hun and Lee, Kunwoo},
  booktitle = {Proceedings of the sixth ACM symposium on Solid modeling and applications},
  pages     = {159--170},
  year      = {2001}
}

@article{lee2023brepgat,
  title     = {BRepGAT: Graph neural network to segment machining feature faces in a B-rep model},
  author    = {Lee, Jinwon and Yeo, Changmo and Cheon, Sang-Uk and Park, Jun Hwan and Mun, Duhwan},
  journal   = {Journal of Computational Design and Engineering},
  volume    = {10},
  number    = {6},
  pages     = {2384--2400},
  year      = {2023},
  publisher = {Oxford University Press}
}

@article{li2012computing,
  title     = {Computing axes of rotation for setup planning using visibility of polyhedral computer-aided design models},
  author    = {Li, Ye and Frank, Matthew C},
  journal   = {Journal of manufacturing science and engineering},
  volume    = {134},
  number    = {4},
  pages     = {041005},
  year      = {2012},
  publisher = {American Society of Mechanical Engineers}
}

@inproceedings{li2021closer,
  title     = {A closer look at rotation-invariant deep point cloud analysis},
  author    = {Li, Feiran and Fujiwara, Kent and Okura, Fumio and Matsushita, Yasuyuki},
  booktitle = {Proceedings of the IEEE/CVF International Conference on Computer Vision},
  pages     = {16218--16227},
  year      = {2021}
}

@article{li2021rotation,
  title     = {A rotation-invariant framework for deep point cloud analysis},
  author    = {Li, Xianzhi and Li, Ruihui and Chen, Guangyong and Fu, Chi-Wing and Cohen-Or, Daniel and Heng, Pheng-Ann},
  journal   = {IEEE transactions on visualization and computer graphics},
  volume    = {28},
  number    = {12},
  pages     = {4503--4514},
  year      = {2021},
  publisher = {IEEE}
}

@inproceedings{li2025dtgbrepgen,
  title     = {DTGBrepGen: A Novel B-rep Generative Model through Decoupling Topology and Geometry},
  author    = {Li, Jing and Fu, Yihang and Chen, Falai},
  booktitle = {Proceedings of the Computer Vision and Pattern Recognition Conference},
  pages     = {21438--21447},
  year      = {2025}
}

@inproceedings{li2025mash,
  title     = {MASH: Masked Anchored SpHerical Distances for 3D Shape Representation and Generation},
  author    = {Li, Changhao and Xin, Yu and Zhou, Xiaowei and Shamir, Ariel and Zhang, Hao and Liu, Ligang and Hu, Ruizhen},
  booktitle = {Proceedings of the Special Interest Group on Computer Graphics and Interactive Techniques Conference Conference Papers},
  pages     = {1--11},
  year      = {2025}
}

@article{liu2025hola,
  title     = {Hola: B-rep generation using a holistic latent representation},
  author    = {Liu, Yilin and Xu, Duoteng and Yu, Xingyao and Xu, Xiang and Cohen-Or, Daniel and Zhang, Hao and Huang, Hui},
  journal   = {ACM Transactions on Graphics (TOG)},
  volume    = {44},
  number    = {4},
  pages     = {1--25},
  year      = {2025},
  publisher = {ACM New York, NY, USA}
}

@article{liu2025manufacturing,
  title     = {Manufacturing process identification from 3D point cloud models using semantic segmentation},
  author    = {Liu, Xiaofang and Wang, Zhichao and Melkote, Shreyes N and Rosen, David W},
  journal   = {Journal of Manufacturing Systems},
  volume    = {82},
  pages     = {858--873},
  year      = {2025},
  publisher = {Elsevier}
}

@inproceedings{lou2023brep,
  title     = {BRep-BERT: Pre-training boundary representation BERT with sub-graph node contrastive learning},
  author    = {Lou, Yunzhong and Li, Xueyang and Chen, Haotian and Zhou, Xiangdong},
  booktitle = {Proceedings of the 32nd ACM International Conference on Information and Knowledge Management},
  pages     = {1657--1666},
  year      = {2023}
}

@article{mandelli2022cad,
  title   = {CAD 3D Model classification by Graph Neural Networks: A new approach based on STEP format},
  author  = {Mandelli, Lorenzo and Berretti, Stefano},
  journal = {arXiv preprint arXiv:2210.16815},
  year    = {2022}
}

@inproceedings{ochmann2019automatic1,
  title        = {Automatic normal orientation in point clouds of building interiors},
  author       = {Ochmann, Sebastian and Klein, Reinhard},
  booktitle    = {Computer Graphics International Conference},
  pages        = {556--563},
  year         = {2019},
  organization = {Springer}
}

@article{ochmann2019automatic2,
  title     = {Automatic reconstruction of fully volumetric 3D building models from oriented point clouds},
  author    = {Ochmann, Sebastian and Vock, Richard and Klein, Reinhard},
  journal   = {ISPRS journal of photogrammetry and remote sensing},
  volume    = {151},
  pages     = {251--262},
  year      = {2019},
  publisher = {Elsevier}
}

@inproceedings{paschalidou2018raynet,
  title     = {Raynet: Learning volumetric 3d reconstruction with ray potentials},
  author    = {Paschalidou, Despoina and Ulusoy, Osman and Schmitt, Carolin and Van Gool, Luc and Geiger, Andreas},
  booktitle = {Proceedings of the IEEE Conference on Computer Vision and Pattern Recognition},
  pages     = {3897--3906},
  year      = {2018}
}

@misc{paviot2022pythonocc,
  author    = {Thomas Paviot},
  title     = {pythonocc},
  version   = {7.7.0},
  year      = {2022},
  publisher = {Zenodo},
  doi       = {10.5281/zenodo.7471333}
}

@inproceedings{qi2017pointnet,
  title     = {Pointnet: Deep learning on point sets for 3d classification and segmentation},
  author    = {Qi, Charles R and Su, Hao and Mo, Kaichun and Guibas, Leonidas J},
  booktitle = {Proceedings of the IEEE conference on computer vision and pattern recognition},
  pages     = {652--660},
  year      = {2017}
}

@article{roth1982ray,
  title     = {Ray casting for modeling solids},
  author    = {Roth, Scott D},
  journal   = {Computer graphics and image processing},
  volume    = {18},
  number    = {2},
  pages     = {109--144},
  year      = {1982},
  publisher = {Elsevier}
}

@inproceedings{shakibajahromi2024rimeshgnn,
  title     = {Rimeshgnn: A rotation-invariant graph neural network for mesh classification},
  author    = {Shakibajahromi, Bahareh and Kim, Edward and Breen, David E},
  booktitle = {Proceedings of the IEEE/CVF Winter Conference on Applications of Computer Vision},
  pages     = {3150--3160},
  year      = {2024}
}

@incollection{shalom2010cone,
  title     = {Cone carving for surface reconstruction},
  author    = {Shalom, Shy and Shamir, Ariel and Zhang, Hao and Cohen-Or, Daniel},
  booktitle = {ACM SIGGRAPH Asia 2010 papers},
  pages     = {1--10},
  year      = {2010}
}

@incollection{stjepandic2015intellectual,
  title     = {Intellectual property protection},
  author    = {Stjepandi{\'c}, Josip and Liese, Harald and Trappey, Amy JC},
  booktitle = {Concurrent engineering in the 21st century: Foundations, developments and challenges},
  pages     = {521--551},
  year      = {2015},
  publisher = {Springer}
}

@inproceedings{su2015multi,
  title     = {Multi-view convolutional neural networks for 3d shape recognition},
  author    = {Su, Hang and Maji, Subhransu and Kalogerakis, Evangelos and Learned-Miller, Erik},
  booktitle = {Proceedings of the IEEE international conference on computer vision},
  pages     = {945--953},
  year      = {2015}
}

@article{takayama2014simple,
  title    = {A simple method for correcting facet orientations in polygon meshes based on ray casting},
  author   = {Takayama, Kenshi and Jacobson, Alec and Kavan, Ladislav and Sorkine-Hornung, Olga},
  journal  = {Journal of Computer Graphics Techniques Vol},
  volume   = {3},
  number   = {3},
  year     = {2014}
}

@inproceedings{tombari2010unique,
  title        = {Unique signatures of histograms for local surface description},
  author       = {Tombari, Federico and Salti, Samuele and Di Stefano, Luigi},
  booktitle    = {European conference on computer vision},
  pages        = {356--369},
  year         = {2010},
  organization = {Springer}
}

@article{velivckovic2017graph,
  title   = {Graph attention networks},
  author  = {Veli{\v{c}}kovi{\'c}, Petar and Cucurull, Guillem and Casanova, Arantxa and Romero, Adriana and Lio, Pietro and Bengio, Yoshua},
  journal = {arXiv preprint arXiv:1710.10903},
  year    = {2017}
}

@article{wang2023manufacturing,
  title     = {Manufacturing process classification based on heat kernel signature and convolutional neural networks},
  author    = {Wang, Zhichao and Rosen, David},
  journal   = {Journal of Intelligent Manufacturing},
  volume    = {34},
  number    = {8},
  pages     = {3389--3411},
  year      = {2023},
  publisher = {Springer}
}

@book{weiler1986topological,
  title     = {Topological structures for geometric modeling (Boundary representation, manifold, radial edge structure)},
  author    = {Weiler, Kevin J},
  year      = {1986},
  publisher = {Rensselaer Polytechnic Institute}
}

@article{weiler2019general,
  title   = {General e (2)-equivariant steerable cnns},
  author  = {Weiler, Maurice and Cesa, Gabriele},
  journal = {Advances in neural information processing systems},
  volume  = {32},
  year    = {2019}
}

@article{willis2021fusion,
  title     = {Fusion 360 gallery: A dataset and environment for programmatic cad construction from human design sequences},
  author    = {Willis, Karl DD and Pu, Yewen and Luo, Jieliang and Chu, Hang and Du, Tao and Lambourne, Joseph G and Solar-Lezama, Armando and Matusik, Wojciech},
  journal   = {ACM Transactions on Graphics (TOG)},
  volume    = {40},
  number    = {4},
  pages     = {1--24},
  year      = {2021},
  publisher = {ACM New York, NY, USA}
}

@inproceedings{wu20153d,
  title     = {3d shapenets: A deep representation for volumetric shapes},
  author    = {Wu, Zhirong and Song, Shuran and Khosla, Aditya and Yu, Fisher and Zhang, Linguang and Tang, Xiaoou and Xiao, Jianxiong},
  booktitle = {Proceedings of the IEEE conference on computer vision and pattern recognition},
  pages     = {1912--1920},
  year      = {2015}
}

@inproceedings{wu2021deepcad,
  title     = {Deepcad: A deep generative network for computer-aided design models},
  author    = {Wu, Rundi and Xiao, Chang and Zheng, Changxi},
  booktitle = {Proceedings of the IEEE/CVF International Conference on Computer Vision},
  pages     = {6772--6782},
  year      = {2021}
}

@article{wu2024aagnet,
  title     = {AAGNet: A graph neural network towards multi-task machining feature recognition},
  author    = {Wu, Hongjin and Lei, Ruoshan and Peng, Yibing and Gao, Liang},
  journal   = {Robotics and Computer-Integrated Manufacturing},
  volume    = {86},
  pages     = {102661},
  year      = {2024},
  publisher = {Elsevier}
}

@article{xu2024brepgen,
  title     = {Brepgen: A b-rep generative diffusion model with structured latent geometry},
  author    = {Xu, Xiang and Lambourne, Joseph and Jayaraman, Pradeep and Wang, Zhengqing and Willis, Karl and Furukawa, Yasutaka},
  journal   = {ACM Transactions on Graphics (TOG)},
  volume    = {43},
  number    = {4},
  pages     = {1--14},
  year      = {2024},
  publisher = {ACM New York, NY, USA}
}

@inproceedings{yu2023rotation,
  title     = {Rotation-invariant transformer for point cloud matching},
  author    = {Yu, Hao and Qin, Zheng and Hou, Ji and Saleh, Mahdi and Li, Dongsheng and Busam, Benjamin and Ilic, Slobodan},
  booktitle = {Proceedings of the IEEE/CVF conference on computer vision and pattern recognition},
  pages     = {5384--5393},
  year      = {2023}
}

@inproceedings{zhang2019rotation,
  title        = {Rotation invariant convolutions for 3d point clouds deep learning},
  author       = {Zhang, Zhiyuan and Hua, Binh-Son and Rosen, David W and Yeung, Sai-Kit},
  booktitle    = {2019 International conference on 3d vision (3DV)},
  pages        = {204--213},
  year         = {2019},
  organization = {IEEE}
}

@inproceedings{zhang2020global,
  title        = {Global context aware convolutions for 3d point cloud understanding},
  author       = {Zhang, Zhiyuan and Hua, Binh-Son and Chen, Wei and Tian, Yibin and Yeung, Sai-Kit},
  booktitle    = {2020 international conference on 3D vision (3DV)},
  pages        = {210--219},
  year         = {2020},
  organization = {IEEE}
}

@article{zhang2022riconv,
  title   = {Riconv++: Effective rotation invariant convolutions for 3d point clouds deep learning},
  author  = {Zhang, Zhiyuan and Hua, Binh-Son and Yeung, Sai-Kit},
  journal = {arXiv preprint arXiv:2202.13094},
  year    = {2022}
}

@article{zhang2024brepmfr,
  title     = {BrepMFR: Enhancing machining feature recognition in B-rep models through deep learning and domain adaptation},
  author    = {Zhang, Shuming and Guan, Zhidong and Jiang, Hao and Wang, Xiaodong and Tan, Pingan},
  journal   = {Computer Aided Geometric Design},
  volume    = {111},
  pages     = {102318},
  year      = {2024},
  publisher = {Elsevier}
}

@article{zhou2008visibility,
  title     = {Visibility-driven mesh analysis and visualization through graph cuts},
  author    = {Zhou, Kaichi and Zhang, Eugene and Wonka, Peter and others},
  journal   = {IEEE Transactions on Visualization and Computer Graphics},
  volume    = {14},
  number    = {6},
  pages     = {1667--1674},
  year      = {2008},
  publisher = {IEEE}
}

@article{zou2025bringing,
  title     = {Bringing attention to CAD: Boundary representation learning via transformer},
  author    = {Zou, Qiang and Zhu, Lizhen},
  journal   = {Computer-Aided Design},
  pages     = {103940},
  year      = {2025},
  publisher = {Elsevier}
}
}

\end{document}